\documentclass[sigconf]{aamas}

\usepackage{hyperref}       
\usepackage{url}            

\usepackage{algorithm}
\usepackage{algpseudocode}
\usepackage{array}
\usepackage{balance}
\usepackage[capitalise,noabbrev]{cleveref}
\usepackage{colortbl}
\usepackage{csquotes}
\usepackage{enumitem}
\usepackage{listings}
\usepackage{longtable}
\usepackage{microtype}      
\usepackage{multirow}
\usepackage{nicefrac}
\usepackage{subcaption}
\usepackage{tabularx}

\setcopyright{ifaamas}
\acmConference[]{Preprint}{}{}{}
\copyrightyear{2026}
\acmYear{2026}
\acmDOI{}
\acmPrice{}
\acmISBN{}

\acmSubmissionID{501}

\title[Agentic Explanations via Interrogative Simulations]{Integrating Counterfactual Simulations with Language Models for Explaining Multi-Agent Behaviour}

\author{Balint Gyevnar}
\orcid{0000-0003-0630-4315}
\affiliation{
  \institution{School of Informatics \\ University of Edinburgh}
  \city{Edinburgh}
  \country{United Kingdom}
}
\email{balint.gyevnar@ed.ac.uk}

\author{Christopher G. Lucas}
\affiliation{
  \institution{School of Informatics \\ University of Edinburgh}
  \city{Edinburgh}
  \country{United Kingdom}
}
\email{c.lucas@ed.ac.uk}

\author{Stefano V. Albrecht}
\affiliation{
  \institution{DeepFlow}
  \city{London}
  \country{United Kingdom}
}
\email{stefano.albrecht@deepflow.com}

\author{Shay B. Cohen}
\affiliation{
  \institution{School of Informatics \\ University of Edinburgh}
  \city{Edinburgh}
  \country{United Kingdom}
}
\email{scohen@inf.ed.ac.uk}

\begin{abstract}
Autonomous multi-agent systems (MAS) are useful for automating complex tasks but raise trust concerns due to risks such as miscoordination or goal misalignment. Explainability is vital for users' trust calibration, but explainable MAS face challenges due to complex environments, the human factor, and non-standardised evaluation. Leveraging the counterfactual effect size model and LLMs, we propose \textit{\textbf{A}gentic e\textbf{X}planations via \textbf{I}nterrogative \textbf{S}imulation (AXIS)}. AXIS generates human-centred action explanations for multi-agent policies by having an LLM interrogate an environment simulator using prompts like \cmd{whatif} and \cmd{remove} to observe and synthesise counterfactual information over multiple rounds. We evaluate AXIS on autonomous driving across ten scenarios for five LLMs with a comprehensive methodology combining robustness, subjective preference, correctness, and goal/action prediction with an external LLM as evaluator. Compared to baselines, AXIS improves perceived explanation correctness by at least 7.7\% across all models and goal prediction accuracy by 23\% for four models, with comparable action prediction accuracy, achieving the highest scores overall. Our code is open-sourced at \href{https://github.com/gyevnarb/axis}{\texttt{https://github.com/gyevnarb/axis}}.
\end{abstract}


\DeclareMathOperator*{\argmax}{arg\,max}

\algnewcommand{\LineComment}[1]{\Statex \textcolor{gray}{#1}}

\definecolor{usercolor}{RGB}{0,100,0}    %
\definecolor{cmdcolor}{RGB}{33, 114, 197}  %
\newcommand{\cmd}[1]{\textcolor{red!70!black}{\texttt{#1}}}

\definecolor{mygreenua}{HTML}{F1F5EB}
\definecolor{myredda}{HTML}{FFE6E6}

\newcommand{\uahelper}[1]{\colorbox{myredda}{${}_{#1\uparrow}$}}
\newcommand{\dahelper}[1]{\colorbox{mygreenua}{${}_{#1\downarrow}$}}
\newcommand{\uaghelper}[1]{\colorbox{mygreenua}{${}_{#1\uparrow}$}}
\newcommand{\dabhelper}[1]{\colorbox{myredda}{${}_{#1\downarrow}$}}

\newcommand{\ua}[1]{\ifthenelse{\equal{#1}{0.00} \or \equal{#1}{0.000}}{}{\uahelper{#1}}}
\newcommand{\da}[1]{\ifthenelse{\equal{#1}{0.00} \or \equal{#1}{0.000}}{}{\dahelper{#1}}}
\newcommand{\uag}[1]{\ifthenelse{\equal{#1}{0.00} \or \equal{#1}{0.000}}{}{\uaghelper{#1}}}
\newcommand{\dab}[1]{\ifthenelse{\equal{#1}{0.00} \or \equal{#1}{0.000}}{}{\dabhelper{#1}}}

\begin{document}

\pagestyle{fancy}
\fancyhead{}

\maketitle

\section{Introduction}\label{sec:intro}

\begin{figure*}
    \centering
    \includegraphics[width=\linewidth]{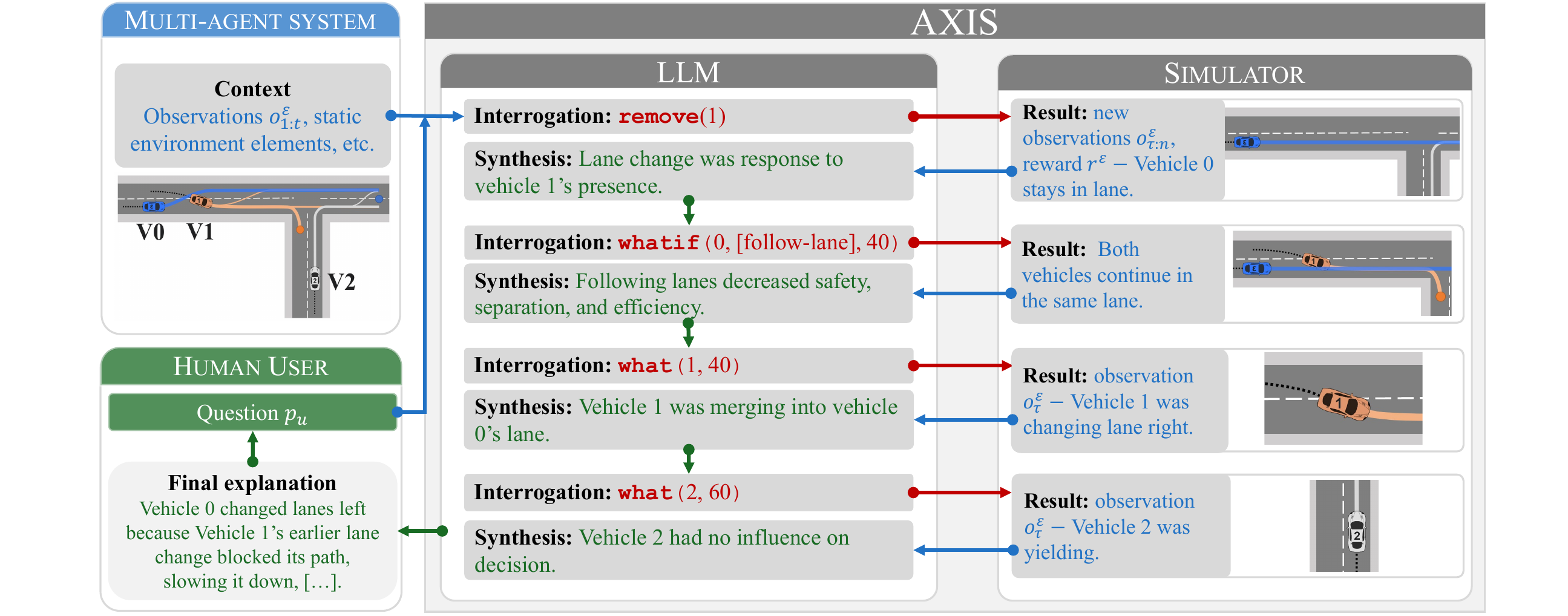}
    \caption{When a user asks a question about an agent's actions, AXIS retrieves the current context, which the LLM uses to interrogate a simulator of the environment, eliciting counterfactual information about the actions of agents.
    The LLM synthesises these counterfactuals into explanations, repeating the process over multiple rounds.
    At the end, AXIS prompts the LLM to synthesise all counterfactuals into a final explanation, which is returned to the user as the answer to their question.
    \textit{Arrow legend:} \textcolor{cmdcolor}{\textit{blue}} pass symbolic data with conversion to text, \textcolor{usercolor}{\textit{green}} pass explanations, \textcolor{red}{\textit{red}} pass interrogation prompts.}
    \label{fig:method}
    \Description{Flowchart of AXIS showing the multi-agent system and the human user on the left with arrows pointing to a box representing AXIS on the right. The box of AXIS has two components, the LLM and the simulator. The LLM has boxes representing the interrogation and synthesis steps. The simulator has boxes representing the results of the interrogation prompts. Arrows connect the interrogation prompts to simulator results, which are then connected to explanation synthesis.}
\end{figure*}

Autonomous agents in multi-agent systems (MAS), e.g. self-driving cars and trading algorithms, are increasingly capable.
They also contribute to the AI-fuelled erosion of human autonomy and oversight, as exemplified by fatal self-driving car incidents~\cite{kalra2016DrivingSafety}, multiple stock market flash crashes~\cite{kirilenko2017flash}, or social media echo chambers~\cite{geschke2019tripleFilterBubble}.
More recently, concerns over emergent risks from advanced artificial intelligence (AI) in MAS, for example collusion, miscoordination, and goal misalignment, have also gained attention~\cite{anthropic2025misalignment}.
Given the above, the trustworthiness, reliability, and uptake of these new technologies is called into question~\citep{hammond2025caif,anthropic2025misalignment}, which is increasingly fuelling stronger and more urgent calls for AI governance~\citep[e.g.][]{iaseai2025statement,coe2024frameworkai}.

A core principle of AI governance is transparency, as facilitated by systems' ability to be explainable~\cite{kaminski2021right, lepri2018fair, wachter2017counterfactual, gyevnar2023transparencyGap}.
Explanations are useful for showing regulatory compliance, because there is evidence to suggest that they help with calibrating people's trust~\cite{dzindolet2003automationtrust, nourani2019meaningfulexpl, yin2019trustaccuracy}, debugging faulty systems~\cite{huber2021highlightsdiv,towers2024temporalxrl,winikoff2017debugging, bewley2021tripletree}, and facilitating effective human-agent teaming~\cite{kim2023helpai, miller2023xaidead, gyevnar2024attribute}.
For autonomous agents, the field of explainable sequential decision-making (SDM) has been motivated by these benefits to develop algorithms for decision-making and control tasks, with most methods focusing on post-hoc feature importance attribution (for surveys, see~\cite{vouros2022xrlsurvey,milani2024xrlsurvey}). %

However, there are at least three roadblocks that hinder progress in the current literature on explainable SDM.
\textbf{(1) Complexity of environments}~\cite{vouros2022xrlsurvey,puiutta2020explainable,milani2024xrlsurvey}, which includes high-dimensional state/action spaces, partial observability, non-stationary dynamics, and (in a multi-agent setting) coupled decision-making, complicating the disentanglement of causal relationships between inputs and actions.
\textbf{(2) The human factor}~\cite{ehsan2024WhoXAI,miller2019socialsciences,dazeley2021LevelsExplainable,milani2024xrlsurvey}, which involves considering the varied cognitive capabilities, domain expertise, and explanation preferences of stakeholders.
\textbf{(3) Lack of standardised evaluation}~\cite{lofstrom2022MetaSurveyQuality,mohseni2021xaievalsurvey,hoffman2023metrics,gyevnar2025objective,dazeley2023xrl}, which limits cross-study comparisons and creates a methodological fragmentation where researchers employ disparate definitions of what constitutes a good explanation.

For MAS, our key insight in tackling these roadblocks is that the actions of agents can be explained in terms of behaviour considering the \emph{causal relationships} among agents' actions, beliefs, desires, intentions and outcomes~\citep{rao1995bdi,lombrozo2006structure,byrne2007rational,winikoff2021badcoffee,byrne2023good}.
We refer to these explanations as \emph{action explanations}, which are a complementary approach to the lower-level mechanistic explanations of SDM policies (e.g. saliency maps or interpretable surrogate models).

To create action explanations, we make three observations based on relevant work.
First, given a joint action policy for a MAS, we can intervene on the actions of agents, leading to counterfactual worlds where we learn about the reactions of other agents.
By summarising counterfactual information, we can create useful explanations~\cite{an2024mctsexplain, gajcin2024counterfactualsxrl, gajcin2024semifactualxrl, admoni2025SySLLM, gyevnar2024causal}.
Second, people prefer explanations that align with their cognitive processes~\cite{miller2019socialsciences, miller2023xaidead, kim2023helpai, gyevnar2024attribute}, and contemporary computational models of causal reasoning suggest that people simulate counterfactual worlds when explaining phenomena~\cite{lucas2015improved, quillien2023counterfactual, droop2023extending}.
Third, LLMs excel at summarising information~\cite{zhang2024benchmarking,brown2020LanguageModelsAre, wang2025recursive, bajaj2021LongDocumentSummarization} and were shown useful in acting as an evaluator~\cite{gu2025surveyllmasajudge,debona2024xaievalllm}.

We operationalise these observations in a new framework called \textit{\textbf{A}gentic e\textbf{X}planations via \textbf{I}nterrogative \textbf{S}imulation (AXIS)}, which generates human-centred action explanations of a given multi-agent action policy.
The overview and an example trace of AXIS execution is shown in~\cref{fig:method}.
We evaluate AXIS on the safety-critical task of motion planning for autonomous driving (AD)~\cite{schwarting2018adplan} across five LLMs using ten scenarios, including ones with irrational agent behaviour and partial observability, with up to three user prompts for each scenario.
We propose a comprehensive evaluation methodology for explainable SDM, combining scenarios with qualitative and computational analyses, subjective preference and perceived correctness scores, and objective accuracy on the downstream tasks of goal and action prediction; using an external LLM as evaluator.

In our evaluation, AXIS improved the perceived correctness by at least 7.7\% across all models and goal prediction accuracy by 23\% for four models, correctly focusing on the \cmd{remove} and \cmd{whatif} interrogations which are most suited to understanding causality.
AXIS also improves or gives comparable results on action prediction for four models and subjective preference scores for three models.
In summary our contributions are:
\begin{enumerate}
    \item We formalise the \textbf{problem of action explanation} for MAS described as partially observable stochastic games;
    \item We propose \textbf{AXIS: Agentic Explanations via Interrogative Simulation} for generating action explanations for any MAS based on counterfactual simulations and LLMs;
    \item We design an \textbf{evaluation methodology for explainable SDM} combining scenarios, qualitative correctness analysis, computational robustness to hyperparameters, and measures of subjective quality and objective actionability;
    \item We conduct \textbf{experiments with autonomous driving} with ten scenarios on five LLMs showing consistently higher perceived explanation correctness and improved goal and action prediction accuracy compared to baselines.
\end{enumerate}

\section{Background}\label{sec:background}

We focus on multi-agent sequential decision-making systems, given as partially observable stochastic games (POSG), using the definitions of~\citet{albrecht2024marlBook} and illustrative examples from AD.
We extend POSGs with the notion of goals following~\cite{albrecht2021igp2}.
The notation $s_{a:b}$ is used for the ordered sequence $(s_a,\dots,s_b)$.

Let $\mathcal{I}$ be the set of indexed agents in the system, e.g. vehicles on a road section.
At timestep $t \in \mathbb{N}^+$, each agent $i \in \mathcal{I}$ is in local state $s^i_t \in \mathcal{S}^i$, e.g. position coordinates and velocity, such that the environment is in joint state $s_t \in \mathcal{S}$, where $\mathcal{S} = \times_{i \in \mathcal{I}} \mathcal{S}^i$.
Agent $i$ receives a local observation $o^i_t \in \mathcal{O}^i$, e.g. states and actions of other vehicles, which probabilistically depends on $s_t$ through $p(o^i_t \mid s_t)$, with the joint observation written as $o_t \in \mathcal{O}$.
In reaction to observations, agent $i$ selects an action $a^i_t \in \mathcal{A}^i$, e.g. acceleration and steering,  following its policy $\pi^i(a^i_t|o^i_{1:t})$.
Once all agents have acted, resulting in joint action $a_t \in \mathcal{A}$, the environment transitions to state $s_{t+1} \sim T(\cdot|s_t,a_t)$ and each agent receives an immediate reward $R^i(s_t,a_t,s_{t+1})$, which may be decomposed into a weighted reward component vector $\vec{R}^i(s_t,a_t,s_{t+1})$, e.g. acceleration, presence of collision. %
The system terminates once it reaches a terminal state in $\bar{\mathcal{S}} \subset \mathcal{S}$ or when a maximum number of steps is reached.
Assuming that local observations and local actions are mutually independent, that is $p(o_t|s_t)=\prod_i p(o^i_t|s_t)$ and $\pi(a_t|o_{1:t})=\prod_i \pi^i(a^i_t|o^i_{1:t})$, then the probability of any joint state sequence $p(s_{1:t})$ is given by:%
\begin{equation}
    \label{eq:p_state}
    p(s_{1:t})=\prod^{t-1}_{n=1} \int_\mathcal{O} \int_\mathcal{A} T(s_{n+1}|s_n,a_n) \pi(a_n|o_{1:n}) p(o_n|s_n) \text{d}o_n \text{d}a_n.
\end{equation}

In the goal-extended version of a POSG, agent $i$ is aiming to reach a goal state $g^i \in \mathcal{G}^i \subset \mathcal{S}^i$, defined as any partial local state description, e.g. destination coordinates.
We say that the joint state sequence $s_{1:t}$ reaches $g^i$ and write $\texttt{Reach}(s_{1:t},g^i)$ if $s^i_t \in g^i$ and $\forall n < t:s^i_n \notin g^i$, i.e. the goal is never achieved sooner than $t$.
The goal $g^i$ is assumed to be unobservable to other agents $j \neq i$.
If agent~$i$ reaches $g^i$, we may calculate the agent's cumulative reward as $R^i(s_{1:t})=\sum_{n=1}^{t-1} R^i(s_t,a_t,s_{t+1})$.

\subsection{A Definition of Action Explanation}\label{ssec:problem}

Our goal is to generate \emph{action explanations} for a given POSG.
We formally define here what is meant by this statement, taking a human-centred perspective on explanations~\cite{miller2019socialsciences,ehsan2020hcxai}.

Assume that a user is observing the POSG and asks a \emph{query} $q \in \mathcal{Q} \subseteq \mathcal{A}^*$ about a sequence of actions of some agents, following a distribution $p(q|s_{1:t})$, where $\mathcal{Q}$ is a finite set of possible queries.
For example, the query $q$ may be a sequence of negative acceleration values of a vehicle that suddenly braked.
Our goal is to generate an \emph{explanation} $e \in \mathcal{E}$ for the query $q$, where $\mathcal{E}$ is a finite set of possible explanations with some syntactic (but not semantic) constraints, e.g. natural language sentences or 64x64 px images.
These explanations are generated by an \emph{explanation function} $E \,\colon \mathcal{Q} \times \mathcal{S}^* \mapsto \mathcal{E}$ that maps a query $q$ and a state trajectory $s_{1:t}$ to an explanation $e=E(q,s_{1:t})$.

Our problem is to find an optimal explanation function $\widehat{E}$ which maximizes some measure(s) of the expected ``quality'' of its generated explanations, where ``quality'' is informally defined to include aspects such as fidelity, human preferences, or actionability, and is formally captured by an external reward model $R: \mathcal{Q} \times \mathcal{E} \mapsto \mathbb{R}$ which maps a query $q$ and explanation $e$ to a scalar value $R(e \mid\mid q)$. %

Given the above definitions, the optimal explanation function is obtained by maximizing the expected external reward over trajectories of all lengths for all possible queries:
\begin{equation}\label{eq:problem-def}
    \widehat{E} \in \argmax_{E \in \Delta(E)} \sum^\infty_{t=1} \sum_{q \in Q} \int_{\Omega_t} p(q, s_{1:t}) R(E(q,s_{1:t}) \mid\mid q)\,ds_{1:t},
\end{equation}
where $\Delta(E)$ is the considered search space of explanation processes, $\Omega_t$ is the set of trajectories we are interested in explaining, the distribution $p(q, s_{1:t}) =  p(q |s_{1:t})p(s_{1:t})$ is the joint probability of a query and state trajectory, and $p(s_{1:t})$ is calculated as in~\cref{eq:p_state}.

The definition of a query $q$ allows a user to inquire about the actions of only one agent (by replacing the actions of other agents with the empty set).
In our experiments, we focus on such queries about a single agent $\varepsilon \in \mathcal{I}$, because single-agent queries are often how people would investigate complex systems~\cite{malle1997WhichBehaviors,olaughlin2002HowPeople,halpern2016actual}.
Importantly, the generated explanations may still involve all agents.

Finally, our formulation produces action explanations that are in terms of elements that can be derived from the state trajectory and queries, e.g. rewards or observations.
Our formulation is not immediately suited for giving a mechanistic interpretation of the functional parametrisation of the policies of agents (e.g. saliency maps) for which other approaches may be more appropriate~\cite{milani2024xrlsurvey}.

\section{Agentic Explanations via Interrogative Simulations}\label{sec:method}

We present a novel method, \textit{\textbf{A}gentic e\textbf{X}planations via \textbf{I}nterrogative \textbf{S}imulations} (AXIS), to approximate $\widehat{E}$, giving an overview in~\cref{fig:method} and the full algorithm in~\cref{alg:axis}.
Our appendix with full method details and the open-sourced codebase are located at \href{https://github.com/gyevnarb/axis}{\texttt{https://github.com/gyevnarb/axis}}.

AXIS is based on the empirically validated \textit{counterfactual effect size model} (CESM)~\cite{quillien2023counterfactual}, which posits that a cause C is more likely to be selected for an explanation of an event E if C co-occurs with E under different but \emph{similar} background conditions (i.e. counterfactuals).
For example, if a car crashes into a tree, then ``the driver being drunk'' might be more often selected as an explanation than ``the presence of the tree'', because under similar background conditions (e.g. weather, road type), whenever the driver is drunk the car crashes, and conversely, whenever the driver is not drunk, the crash is absent.
Meanwhile, whenever the tree is not present (but the driver is drunk), a crash may still occur.

AXIS converts the CESM into a computational method by using an LLM to propose interventions on the actions of agents and then forward-simulating the counterfactual system to observe whether the query still occurs.
As a corollary requirement, the main assumption of AXIS is the existence of a simulator $\widehat{T}$ which approximates the state transition function $T$ and the joint action policy $\pi$.
When data is available, a parameterised model of future trajectories could be learned as $\widehat{T}$.
When data is not available, a range of modelling and planning algorithms can be deployed to obtain $\widehat{T}$~\cite{albrecht2018modelling}.

\paragraph{Options}

An important question with using an LLM to propose interventions is whether the representations in a POSG are immediately understandable to the LLM, and vice versa; whether the instructions of an LLM mean anything to the simulator of a POSG.
For example, actions in AD are sequences of real-valued vectors which are not easily parsed by an LLM for longer sequences~\cite{liu2024lostinmiddle}.
In addition, LLM reasoning over low-level actions may not be effective for long-horizon explanations~\cite{liu2024lostinmiddle}, unless $|\mathcal{A}|$ is finite and small, but this is not generally the case.

To bridge any representational gaps between LLM and simulator, we may use action abstractions based on the popular formalism of \emph{options} from~\citet{sutton1999MDPsSemiMDPs}.
Given a finite set of options $\mathcal{M}$, an option $m \in \mathcal{M}$ is a (named) triple $(\mathcal{S}_m,b_m,\pi^i_m)$ where $\mathcal{S}_m \subseteq \mathcal{S}$ is a set of initial states, $b_m\subseteq \mathcal{S}$ is a set of termination states, and $\pi^i_m:\mathcal{S} \times \mathcal{A}^i \mapsto [0,1]$ is the policy used by the option.
Intuitively, an option groups action sequences that begin when an agent reaches an initial state in $\mathcal{S}_m$, then follows the (stochastic) policy $\pi^i_m$ until a termination state in $b_m$ is reached.
In AXIS, the LLM proposes interventions in the form of options, which are then executed by the simulator.
To wrap observed action sequences into options for the LLM, written as $\texttt{wrap}(a^i_{1:t})$, we check the initial and termination states of all options against the sequence until a match is found.

\paragraph{Simulator \& Interrogation Prompts}

\begin{table}
\centering
\small
\caption{Interrogation prompts used to facilitate communication between the LLM and the simulator.
}
\label{tab:prompt_language}
\begin{tabular}{@{}lp{4cm}p{2.5cm}@{}}
\toprule
\textbf{Prompt} & \textbf{Arguments and Effect} & \textbf{Example} \\
\midrule

\cmd{add} &
\textbf{Args:} $\texttt{start} \subset \mathcal{S},\, \texttt{goal} \subset \mathcal{S}$ \newline
\textbf{Effect:} Add agent with initial state \texttt{start} and goal state \texttt{goal}. &
\cmd{add}($[2,68], [-3,14]$) \newline
\textit{“Agent at [2,68] with goal [-3,14].”} \\

\cmd{remove} &
\textbf{Args:} $\texttt{agent} \in \mathcal{I}$ \newline
\textbf{Effect:} Remove specified \texttt{agent} from start of the simulation. &
\cmd{remove}($1$) \newline
\textit{“Remove the agent with ID 1.”} \\

\cmd{whatif} &
\textbf{Args:} $\texttt{agent} \in \mathcal{I}$, $\texttt{option} \in \mathcal{M}^*$, $\texttt{time} \in \mathbb{N}^+$ \newline
\textbf{Effect:} Simulates the \texttt{agent} executing \texttt{option} starting at \texttt{time}. &
\cmd{whatif}(1 [\cmd{turn}], 40) \newline
\textit{“What if agent 1 turns at 40?”} \\

\cmd{what} &
\textbf{Args:} $\texttt{agent} \in \mathcal{I},\, \texttt{time} \in \mathbb{N}^+$ \newline
\textbf{Effect:} Query \texttt{agent}’s state/action at observed/simulated \texttt{time}. &
\cmd{what}($1$, $40$) \newline
\textit{“What was agent 1 doing at 40?”} \\

\bottomrule
\end{tabular}
\end{table}

Building on options, we facilitate communication between the LLM and the simulator through a set of \textit{interrogation prompts} shown in~\cref{tab:prompt_language}.
These prompts allow the LLM to propose interventions to agents' behaviour and the simulator to interpret those proposals.
An interrogation prompt $p$, such as \textit{`What if vehicle 2 had changed lanes at time 40?'}, consists of a keyword followed by arguments in parentheses with arguments referred to as $p.\texttt{arg}$, e.g. \cmd{whatif}($2, [\texttt{change-lane}], 40$).

With this setup, the simulation process proceeds as follows.
The simulator takes as input the current sequence of observations $o^\varepsilon_{1:t}$ from the perspective of the queried agent $\varepsilon$.
The LLM is then requested to propose an interrogation prompt $p$.
In case $p.\texttt{time} \neq t$, we either roll back time by truncating $o^\varepsilon_{1:t}$ or simulate forward the environment until the starting simulator-time, written as $\tau$, matches $q.\texttt{time}$.
The environment is then forward simulated using $\widehat{T}$, overriding agents' policy with the specified intervention in prompt $p$ as applicable, until the simulation terminates, returning new joint states $s_{\tau:n}$ and actions $a_{\tau:n}$, where $n$ is a variable final time step.

\paragraph{Verbalisation}

We call \textit{verbalisation}, and write \texttt{verbalise}($\cdot$), the process of converting into text the various representations of a POSG for the LLM.
For AXIS, we may verbalise the observation sequences ($V_o$), the options ($V_m$), and the reward signal ($V_r$), as well as optionally the static environment elements and metadata ($V_{env}$),
We refer to these objects as \emph{context features} (or features), because the verbalised objects are used to provide context to the LLM.

To obtain $V_o$, we convert $o^\varepsilon_{1:\tau}$ to the textual representations of the state and action vectors in the observations as provided by our programming language, noting that for complex state spaces or very long sequences, state filtering~\cite{albrecht2016exploiting} or state abstraction~\cite{abel2018stateAbstraction,gyevnar2024causal} may be necessary to produce useful observation verbalisations.
In addition, to get $V_m$, we \texttt{wrap} the joint action sequence $a_{1:\tau}$ to options and use their associated names and start and end time steps.
For verbalising rewards $V_r$, we calculate the cumulative reward $R^\varepsilon(s_{1:n})$, and if available, decompose it to reward components $\vec{R}^i(s_{1:n})$.
An example for each verbalisation element is given in Appendix~A.

Finally, we define five LLM prompt templates, \texttt{system}, \texttt{context}, \texttt{interrogation}, \texttt{explanation}, and \texttt{final}, to dynamically create prompts from verbalisation.
Each template contains a number of placeholders, e.g. for the verbalised observations or the user prompt, which are replaced with the appropriate strings during runtime.
The complete text of AXIS templates is given in Appendix~B.

\paragraph{AXIS Algorithm}\label{ssec:method:algorithm}

\begin{algorithm}[t]
\caption{The explanation function of AXIS. Note, whenever the LLM is prompted, the prompt is appended to a prompt history.}
\label{alg:axis}
\begin{algorithmic}[1]
\Require User question $p_u$, observations $o^\varepsilon_{1:t}$, simulator $\widehat{T}$, maximum rounds $N_{\max}$, optional verbalised static elements $V_{env}$.
\Ensure Natural language action explanation $e$.

\State $H \gets []$ empty prompt history.
\State $(V_o,V_m,\varnothing) \gets$ \textsc{VerbaliseContext}($o^\varepsilon_{1:t},\varnothing$)
\State Fill \texttt{context} template from $V_{env}$, $V_o$, $V_m$, $p_u$ and add to $H$.

\For{$N$ \textbf{in} $[1,\dots,N_{\max}]$}
    \LineComment{\quad\texttt{Interrogation phase}}
    \State Fill \texttt{interrogation} template with $p_u$.
    \State $p \gets$ Prompt LLM for interrogation prompt with $H$.
    \State \textbf{if} {$p==\texttt{DONE}$} \textbf{then break} \Comment{\texttt{Early termination}}
    \State $s_{\tau:n}, a_{\tau:n}, r^\varepsilon \gets$ Run simulation with $\widehat{T}$ and $p$.

    \LineComment{\quad\texttt{Synthesis phase}}
    \State $(V_o, V_m, V_r) \gets$ \textsc{VerbaliseContext}($[s_{\tau:n},a_{\tau:n}],\,r^\varepsilon$)
    \State Fill \texttt{explanation} template with $V_o$, $V_m$, $V_r$, $p_u$, add to $H$.
    \State $e \gets$ Prompt LLM for explanation with $H$.
\EndFor

\State Fill \texttt{final} template from $p_u$.
\State $e \gets$ Prompt LLM for final explanation with $H$.
\State\Return $e$
\Function{VerbaliseContext}{$o^\varepsilon_{1:t}$, $r^\varepsilon$}
    \State $a_{1:t} \gets$ Extract joint actions of visible vehicles from $o^\varepsilon_{1:t}$.
    \State $V_o \gets \texttt{verbalise}(o_{1:t})$
    \State $V_m \gets \texttt{verbalise}(\{ \texttt{wrap}(a^i_{1:t}) \mid i \in I\})$
    \State $V_r \gets \texttt{verbalise}(r^\varepsilon)$ \textbf{if} $r^\varepsilon \neq \varnothing$ \textbf{else} $\varnothing$
    \State \Return $(V_o, V_m, V_r)$
\EndFunction
\end{algorithmic}
\end{algorithm}

The pseudocode of AXIS is given in~\cref{alg:axis}.
AXIS starts when a user prompts it for an action explanation regarding the actions of an agent $\varepsilon$.
We write $p_u$ for the natural language user question, with $t$ denoting the time of prompting and $\varepsilon$ denoting the queried agent.
AXIS then resets its prompt history $H$ and calls the \textsc{VerbaliseContext} function to obtain the verbalised observations and options.
A \texttt{context} template is filled using the verbalised information along with $p_u$ and added to $H$.

The core of~\cref{alg:axis} is a multi-round interrogation and synthesis process that runs for up to $N_{\max}$ rounds.
During each round $N$, an \texttt{interrogation} template is filled based on $p_u$.
The LLM is then prompted to generate an interrogation prompt $p$ based on $H$.
If $p$ is \texttt{DONE}, the loop terminates, which allows the LLM to stop the action explanation generation process early.
Otherwise, a simulation is run with $\widehat{T}$ based on the interrogation query, producing new states $s_{\tau:n}$, actions $a_{\tau:n}$, and rewards $r^\varepsilon = R^\varepsilon(s_{\tau:n})$.

Following interrogation, the explanation synthesis phase begins.
The \textsc{VerbaliseContext} function is called again, this time including rewards $r^\varepsilon$. %
An \texttt{explanation} template is filled with the verbalised components and $p_u$, and added to $H$.
The LLM then generates an explanation $e$ based on the updated prompt history.
After completing all rounds, a \texttt{final} prompt template is filled based on $p_u$, and the LLM is prompted with the complete prompt history for a final action explanation $e$, which is returned as the algorithm's output.

\section{Evaluation Methodology}\label{sec:evaluation}

We evaluate AXIS on the task of motion planning for AD~\cite{schwarting2018adplan}, a safety-critical decision-making problem, where explanations may be particularly useful for safety analysis and calibrating trust~\citep{kuznietsov2024avreview}.
The simulator $\widehat{T}$ we use for this purpose is based on a method called IGP2 by~\citet{albrecht2021igp2}, which operates in a 2D environment on a static semantically annotated road layout with vehicles following either A* search or Monte Carlo Tree Search planning algorithms.

Unfortunately, the evaluation of explanations is notoriously difficult~\cite{milani2024xrlsurvey,miller2023xaidead,mohseni2021xaievalsurvey,rudin2019stopexplaining}.
The perceived utility of an explanation depends not only on its causal content but also on the context and background of people interpreting the explanation~\cite{ehsan2024WhoXAI,miller2019socialsciences,gyevnar2024attribute}.
To mitigate these issues, we use scenario-based evaluation with a novel combination of subjective metrics of user preferences and objective metrics of actionability for goal and next action prediction.

\begin{figure}
    \centering
    \includegraphics[width=0.49\linewidth]{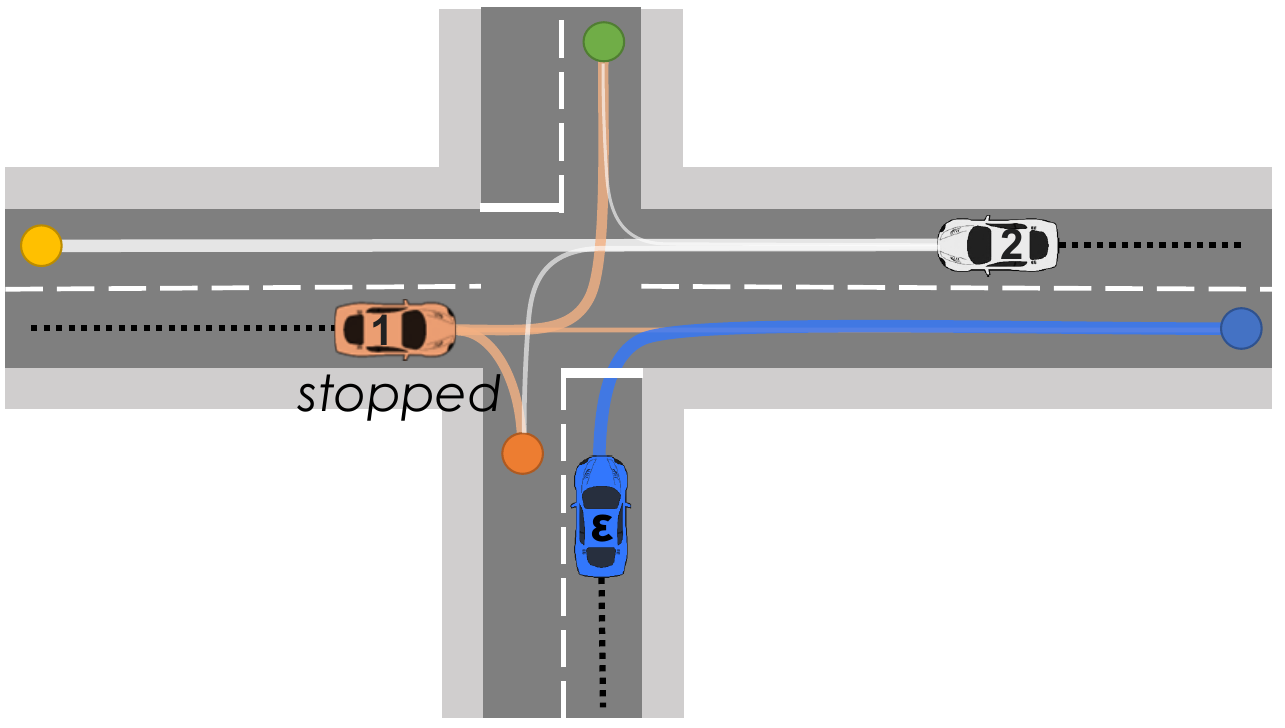}
    \includegraphics[width=0.49\linewidth]{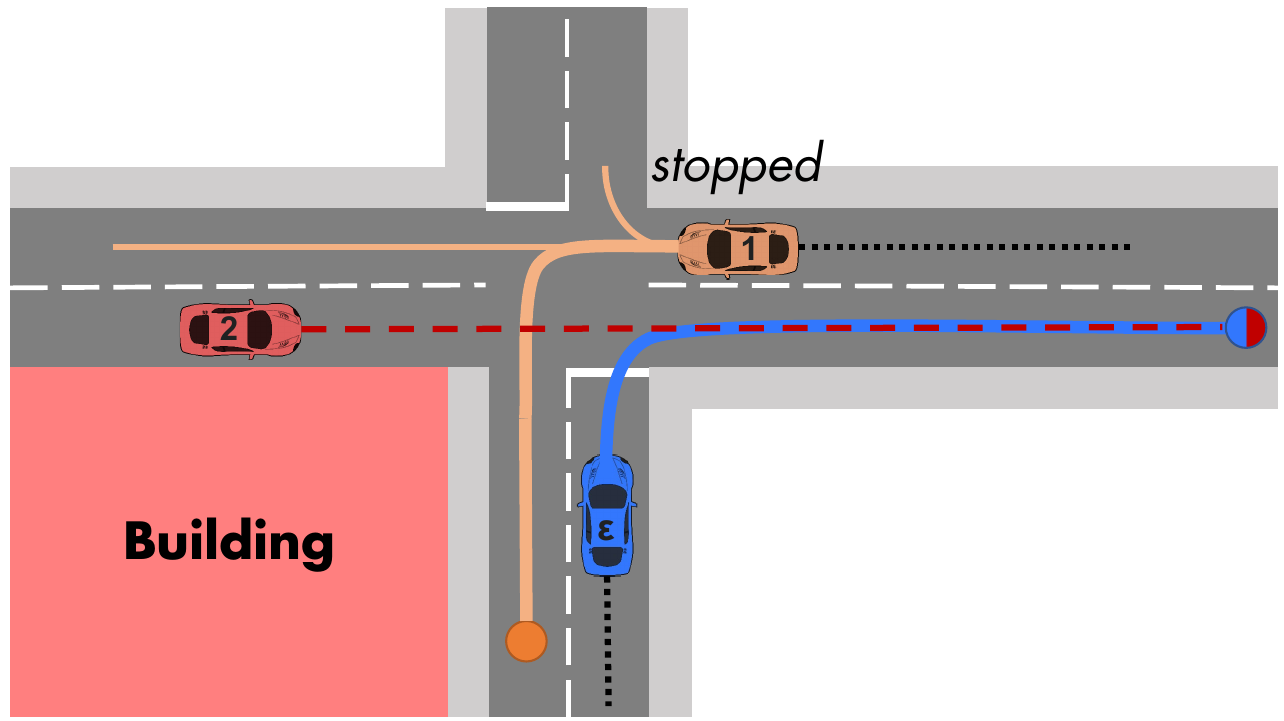}
    \caption{Example scenarios with the \textcolor{blue}{queried vehicle} shown in blue, \textcolor{orange}{other vehicles} in orange/white, and \textcolor{red}{occluded vehicles} in red. Scenario \#3 (left; rational): blue sees orange come to a stop, indicating orange's intent to turn left, so blue decides to turn right instead of waiting longer to yield. Scenario \#8 (right; occlusion): blue sees orange on a priority lane coming to a stop, inferring that red must be behind the building on the left, so blue stops.}
    \label{fig:scenarios}
    \Description{Three 2D top-down visualisations of driving scenarios depicting the scenarios as explained in the caption.}
\end{figure}

\paragraph{Scenarios \& User Prompts}

We use ten scenarios with up to five agents (i.e. vehicles) based on previous work~\cite{albrecht2021igp2, hanna2021gofi, gyevnar2024attribute} where explanations were shown to be useful for users.
Example scenarios are shown in~\cref{fig:scenarios}, with full descriptions in Appendix~C.
Scenarios capture tactical driving situations, e.g. lane changing, unprotected turn, lane merging, and roundabout entry, and involve either safety or efficiency-related reactions from the agents.
Scenarios have three categories: five have normal, rational driving behaviour (rational scenarios \#1-5), two contain irrational behaviour from a non-ego vehicle (irrational scenarios \#6-7), and three contain occluded vehicles (occlusion scenarios \#8-10).
For each scenario, we define up to three user questions $p_u$, e.g. \textit{`Why did vehicle 0 change lanes left?'} and \textit{`What would have happened if vehicle 1 went straight?'}, and generate explanations for each question separately.

\paragraph{Baselines \& Models}
We compare against two baselines.
\textbf{(1) Model only (\textit{ModelOnly})}, where we give the full initial context to the LLM and ask it to explain the scenario.
\textbf{(2) No explanation (\textit{NoExp})}, only applicable for actionability metrics, where we do not provide any explanation to the evaluator, to establish a baseline capacity to predict the agent's goal and next action.
We are not aware of relevant previous work that is directly comparable to AXIS.

Five LLMs are used to generate explanations and test the robustness of AXIS to the type and size of LLMs: OpenAI GPT-4.1-2025-04-14~\cite{gpt41}, o1-2024-12-17~\cite{o1}, DeepSeek-V3-0324~\cite{deepseekv3}, DeepSeek-R1~\cite{deepseekr1}, and Meta Llama 3.3-70B-Instruct~\cite{llama70b}.
We also experimented with LLMs up to 7B parameters, but the results were inaccurate as the models only copied text from the context for interrogation queries, without actually proposing meaningful interventions.
For sampling from the LLMs, we use default hyperparameters across all models, given in Appendix~D, and set a maximum output token limit of 512 for instruct and 16,384 for reasoning models.
We use $N_{max}=10$ maximum rounds in AXIS.

\paragraph{Metrics}

Unlike previous work in explainable SDM which usually centres on subjective preference and, infrequently, on downstream actionability~\cite{milani2024xrlsurvey}, we use a combination of five different approaches. %
This evaluation setup allows us to qualitatively assess the explanations' correctness and measure their robustness to variation, as well as understand users' subjective preferences, perceived explanation correctness, and downstream performance (\emph{$^*$require user study}):
\begin{enumerate}
    \item \textbf{Qualitative correctness} is assessed through case studies and by identifying which causes are selected for the action explanation of AXIS.
    \item \textbf{Robustness to variations} is measured by varying the experimental conditions, e.g. hyperparameters and LLM models, to see whether correct explanations are generated even under different conditions.
    \item \textbf{Subjective preference$^*$} is measured along four 5-point Likert scales following~\citet{hoffman2023metrics}: perceived completeness, satisfaction, sufficient detail, and trustworthiness of the explanation. We calculate the geometric mean of the four scales and report a single score between 1 and 5.
    \item \textbf{Perceived correctness$^*$} is calculated by comparing an expert causal explanation of the scenario, similar to those in~\cref{fig:scenarios}, with full description in Appendix~D, to an explanation by AXIS, reported on a scale between 1, completely incorrect, to 5, completely correct.
    \item \textbf{Goal and next action prediction$^*$} measures the accuracy of people predicting the goal and immediate next action of agents given an explanation based on~\citet{gyevnar2025objective}, reported as accuracy in the interval $[0,1]$.
\end{enumerate}
When presenting aggregate results, we select best-round explanations which have the highest geometric mean of preference and perceived correctness scores across interrogation-synthesis rounds.

\paragraph{LLM-as-a-Judge}
User studies are expensive and hard to control. %
In order to scale our evaluation setup, we use the popular approach of LLM-as-a-judge~\citep{gu2025surveyllmasajudge}.
This is possible, because AXIS is targeted at end users in an accessible domain such as driving for which ample training data may be available.
Previous work also showed that LLM-based substitutes of crowdsourced human data for explanation evaluation are feasible~\citep{debona2024xaievalllm}, although there may be limitations of this setup, which we will discuss in detail in~\cref{sec:results,sec:conclusion}.

For our LLM-as-a-judge approach, we design prompt templates for each user-study-based metric (metrics 3 to 5 above) and use Anthropic Claude 3.5~\cite{anthropic2025claude}, an LLM not used for explanation generation, as evaluator. %
Full prompt templates used for evaluation are in Appendix~D.
We manipulate three independent within-subjects variables in this experimental setup: \textbf{(1) Model}: the LLM model used in AXIS, \textbf{(2) Scenario}: the scenario explained by AXIS, and \textbf{(3) Query}: the natural language user question as listed in Appendix~C.

\paragraph{Shapley Analysis of Context Features}\label{ssec:eval:shapley}

\begin{figure}
    \centering
    \begin{subfigure}[b]{0.9\linewidth}
        \centering
        \includegraphics[width=\textwidth]{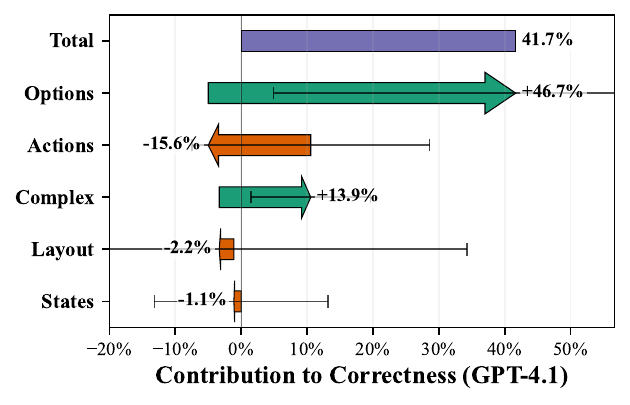}
        \label{fig:shapley_gpt41}
    \end{subfigure}
    \hfill
    \begin{subfigure}[b]{0.9\linewidth}
        \centering
        \includegraphics[width=\textwidth]{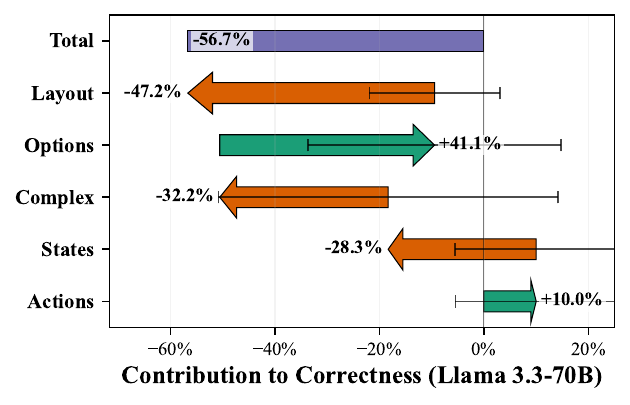}
        \label{fig:shapley_llama70b}
    \end{subfigure}
    \caption{Shapley values calculated from perceived correctness for GPT-4.1 (top) and Llama 3.3-70B (bottom) models. The $x$-axis shows the Shapley value contribution to the final correctness score. The blue Total bar shows the cumulative Shapley value with all features. Results are aggregated across scenarios \#3, \#7, \#8. Error bars show standard error of mean.}
    \label{fig:shapley}
    \Description{Two bar charts showing positive Shapley values in green and negative Shapley values in orange. The top chart for GPT-4.1 shows green bars for Options and Complex, and orange bars for Actions, Layout, and States. The bottom chart for Llama 70B shows green bars for Options and Actions, and orange bars for Layout, States, and Complex.}
\end{figure}

We use Shapley value analysis~\citep{shapley1953value} for context feature selection based on perceived correctness to optimise explanation correctness and to shorten prompt lengths.
Due to the factorial number of evaluations required for a complete Shapley analysis, we evaluate on a subset of models and scenarios, while covering a broad range of experimental conditions.
We select two LLMs, one smaller (Llama 3.3-70B) and one larger (GPT-4.1), to compare the differences in model size.
For the two LLMs, we choose three scenarios (\#3, \#7, \#8), each covering one of the three scenario types (rational, irrational, occlusion).
For the 2 LLMs and 3 scenario types, we test five AD-specific context features which involved a total of $2\times 3\times 5!=720$ runs of AXIS:
\begin{itemize}
    \item \textbf{Layout} ($V_{env}$): The semantic road layout, driving rules, and metadata (e.g. units, speed limit) described in text.
    \item \textbf{States} ($V_o$): The states contained in the observations, i.e. position and velocity vectors, and scalar heading.
    \item \textbf{Actions} ($V_o$): The actions contained in the observations, i.e. acceleration and steering vectors.
    \item \textbf{Options} ($V_m$): The high-level options based on IGP2~\cite{albrecht2021igp2}, i.e. \texttt{Continue}, \texttt{Change}-\{\texttt{Left},\texttt{Right}\}, \texttt{Exit}, and \texttt{Stop}
    \item \textbf{Linguistic Complexity}: Additional feature controlling linguistic complexity of the explanation, i.e. concise or complex.
\end{itemize}

We select features for our full experiments based on the Shapley analysis, with aggregate results over scenarios shown in~\cref{fig:shapley} and per-scenario results in Appendix~E.
Reward components were always included in the prompt as they are essential for explanations in terms of the desires and goals.
We find that options are consistently helpful for both models, with the larger model using state and layout information more effectively, and the smaller model using action information more effectively.
The larger model also benefits from linguistic complexity, and the smaller model from concise explanations.
Therefore, we select options, complex linguistic content, and states for larger models, and options, actions, and concise linguistic complexity for the smaller model.

\section{Results}\label{sec:results}

We ask the following two research questions to understand whether AXIS is effective at generating useful action explanations:
\begin{enumerate}
    \renewcommand{\labelenumi}{(RQ\arabic{enumi})}
    \item Does the interrogation and synthesis process help with selecting the correct causes behind the actions of the queried vehicle in each scenario?
    \item Are action explanations of AXIS preferred and perceived correct by users, and do they help users predict the agent's goal and next action?
\end{enumerate}
We assess RQ1 through our qualitative and robustness analyses in and RQ2 through the subjective and actionability metrics.

\subsection{Qualitative Assessment}
\label{sec:qualitative}

\begin{table}
    \centering
    \caption{Verbatim excerpts from AXIS for scenario \#2, given to the user prompt \textit{``Why did vehicle 0 change lanes left?''}}
    \label{tab:qualitative}
    \renewcommand{\arraystretch}{1.1}
    \begin{tabular}{@{}p{\linewidth}@{}}
    \toprule
    \textbf{GPT-4.1 (ModelOnly):} Vehicle 0 changed left because vehicle 1, originally in the adjacent lane, completed a rightward lane change and moved ahead, creating space. With no immediate obstruction and an opportunity to overtake or maintain a higher speed, vehicle 0 initiated the lane change to improve its driving conditions. \\
    \midrule
    \textbf{GPT-4.1:} Vehicle 0 changed lanes left to avoid a \textit{potential conflict with vehicle 1.} [...]
    This maneuver enabled vehicle 0 to \textit{maintain its speed} and \textit{avoid being impeded} or forced to slow down. The \textit{actions and timing of vehicle 1}, rather than vehicle 2, were the primary cause for vehicle 0's lane change. [...] \\

    \textbf{DeepSeek-V3:} Vehicle 0 changed lanes left to avoid a potential \textit{conflict with Vehicle 1}, which was actively merging right into its path. The counterfactual tests showed: 1) When Vehicle 0 stayed in its lane, it had to \textit{slow down significantly} due to Vehicle 1's merge 2) When \textit{Vehicle 1 didn't merge, Vehicle 0 maintained smooth progress} without lane changes. [...] \\

    \textbf{Llama 70B:} Vehicle 0 changed lanes left \textit{in response to Vehicle 1's lane change}, likely to \textit{avoid a potential conflict} or make space, indicating a causal relationship between their actions. \\

    \textbf{o4-mini:} Vehicle 0's left-lane move wasn't a random drift but a deliberate way to \textit{maintain speed and clear the upcoming junction}. In the original run, the car beside it cut in and was going to slow or turn, so changing lanes let Vehicle 0 \textit{avoid braking}. \\

    \bottomrule
    \end{tabular}
\end{table}

Our results show that AXIS extracts the most important causes for rational scenarios.
For example, looking at explanations of scenario~\#2 in~\cref{tab:qualitative}, all models cite correct causes in terms of other agents' actions and the queried agent's rewards, including effects on speed, strategic maneuvering, and safety, while adding useful contextual information, e.g. highlighting irrelevant vehicles and road layout information.
Explanations are often contrastive as well, which are argued to be most intelligible to people~\cite{miller2019socialsciences}.

In comparison, the ModelOnly baseline is generic, not contrastive and fails to identify the most important causes, incorrectly attributing the lane change of the queried vehicle to vehicle 1 \enquote{completing a rightward lane change and moving ahead, creating space.}
This highlights the benefit of AXIS's interrogation and synthesis process in selecting the most relevant causes for the explanation.
Similarly, intelligible and correct explanations are generated by AXIS for scenarios \#1 to \#5, as shown in Appendix~F.

However, AXIS encounters the limitations of LLMs when having to reason about irrational behaviour or when occluded factors are present.
In the case of scenarios with irrational agents, the LLMs mostly ignore the irrational behaviour, focusing entirely on the actions of the queried agent.
The LLMs produce incomplete explanations because they seem to assume all vehicles to obey the driving rules and act rationally, so they never propose interventions where agents are forced to act irrationally.
In the case of occlusion scenarios, the LLMs were explicitly told that occluded vehicles may be present, see Appendix~B, yet they either failed to extract occlusion-related causes or assumed that all non-queried vehicles were occluded. %
These issues are a result of the limitations of LLMs, in spite of which AXIS still manages to improve overall scores, as we show in~\cref{ssec:quantitative}.
It is also worth noting that reasoning about partial observability and irrationality is a very difficult challenge for any decision-making system in general~\cite{liu2022partially,masters2019goal,hanna2021gofi}. %

\begin{figure}
    \centering
    \includegraphics[width=\linewidth]{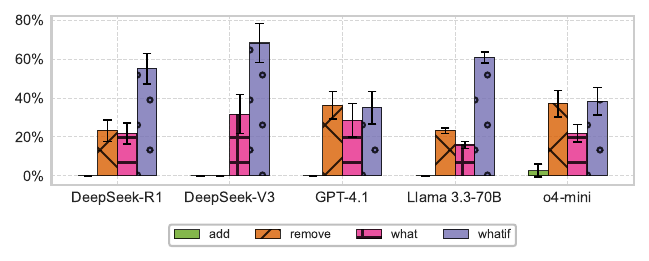}
    \caption{The proportion of queries across all scenarios and prompts used by the different models. Most models focus on \cmd{remove} and \cmd{whatif} queries which are most suited to extract counterfactuals.}
    \label{fig:query_counts}
    \Description{Bar chart with models on the x-axis and proportion of query types on the y-axis, and different colours for different query types. The ``add'' prompt is always lowest, usually followed by the ``what'' prompt, then ``remove'', then ``whatif''.}
\end{figure}

Nevertheless, \cref{fig:query_counts} shows that LLMs do focus on using the \cmd{whatif} and \cmd{remove} prompts across all scenarios.
These prompts are the most useful for obtaining causal information, further highlighting the benefit of the interrogation phase in AXIS.
For example, removing a vehicle can impact the behaviour of surrounding vehicles by allowing them to take simpler actions without the need to adjust to the removed vehicle.
When a model failed to use these prompts, for example with DeepSeek-V3, it achieved the lowest perceived correctness (cf.~\cref{tab:axis:eval_results}), suggesting that it failed to understand the appropriate causal relationships.

\subsection{Quantitative Analysis}\label{ssec:quantitative}

\begin{table*}
\caption{Comparison of baselines and AXIS on metrics aggregated across scenarios and prompts, with AXIS using best-round explanations. Red/green values show the change of mean relative to the model-only baseline. AXIS improves the correctness of explanations for all models and actionability for most models. It also achieves the highest scores across all metrics.}
\label{tab:axis:eval_results}
\centering
\setlength{\tabcolsep}{3pt}  %
\rowcolors{3}{gray!10}{white}

\resizebox{\textwidth}{!}{
\begin{tabular}{@{}llllllllllll@{}}
\toprule
\textbf{Metric} &
\textbf{NoExp} &
\textbf{GPT-4.1} &
\textbf{+AXIS} &
\textbf{DSV3} &
\textbf{+AXIS} &
\textbf{Llama 70B} &
\textbf{+AXIS} &
\textbf{o4-mini} &
\textbf{+AXIS} &
\textbf{DSR1} &
\textbf{+AXIS} \\
\midrule

\textit{Preference} &
N/A &
2.86$\pm$0.14 &
2.50$\pm$0.19\dab{-.36} &
2.69$\pm$0.14 &
\textbf{3.49}$\pm$0.17\uag{+.80} &
2.81$\pm$0.15 &
2.36$\pm$0.15\dab{-.45} &
3.19$\pm$0.20 &
3.15$\pm$0.19\dab{-.04} &
3.33$\pm$0.20 &
3.37$\pm$0.21\uag{+.04} \\

\textit{Correctness} &
N/A &
3.00$\pm$0.27 &
3.23$\pm$0.34\uag{+.23} &
2.64$\pm$0.22 &
3.18$\pm$0.32\uag{+.55} &
3.14$\pm$0.34 &
3.59$\pm$0.29\uag{+.45} &
3.59$\pm$0.33 &
\textbf{4.23}$\pm$0.25\uag{+.64} &
3.73$\pm$0.31 &
4.00$\pm$0.25\uag{+.27} \\

\textit{Goal Acc.} &
0.56$\pm$0.05 &
0.59$\pm$0.11 &
\textbf{0.73}$\pm$0.10\uag{+.14} &
0.50$\pm$0.11 &
\textbf{0.73}$\pm$0.10\uag{+.23} &
0.45$\pm$0.11 &
0.64$\pm$0.10\uag{+.18} &
0.55$\pm$0.11 &
0.68$\pm$0.10\uag{+.14} &
0.50$\pm$0.11 &
0.50$\pm$0.11\uag{0.00} \\

\textit{Action Acc.} &
0.53$\pm$0.05 &
0.59$\pm$0.11 &
\textbf{0.73}$\pm$0.10\uag{+.14} &
0.59$\pm$0.11 &
0.45$\pm$0.11\dab{-.14} &
0.27$\pm$0.10 &
0.64$\pm$0.10\uag{+.36} &
0.59$\pm$0.11 &
0.55$\pm$0.11\dab{-.05} &
0.59$\pm$0.11 &
0.50$\pm$0.11\dab{-.09} \\

\bottomrule
\end{tabular}
}
\vspace{-1ex}
\end{table*}

\begin{figure}
    \centering
    \includegraphics[width=0.81\linewidth]{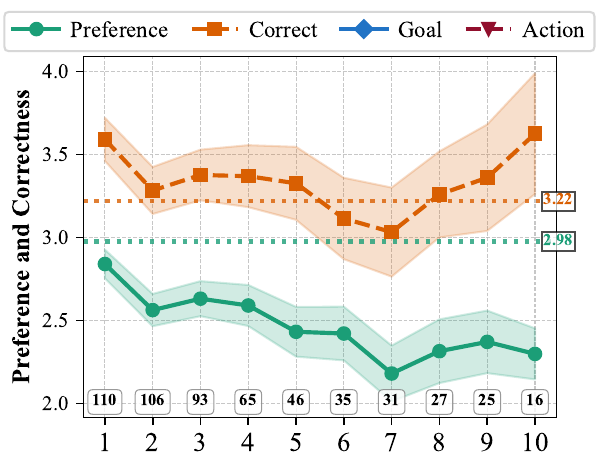}
    \includegraphics[width=0.78\linewidth]{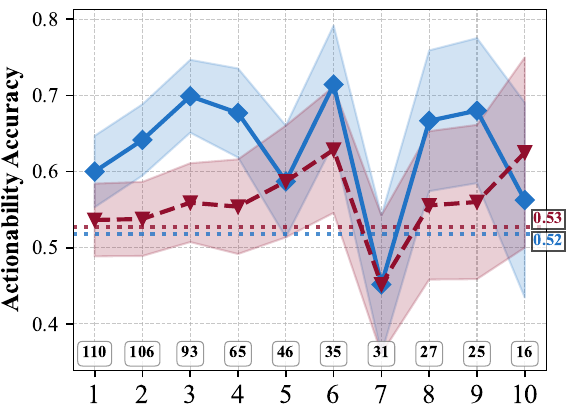}
    \caption{Evolution of preference/correctness (\textbf{top}), and goal/action prediction accuracy (\textbf{bottom}) of intermediate explanations over interrogation-synthesis rounds, aggregated across models, scenarios, and user prompts. The number of explanations for each round is shown above the $x$-axis in boxes. Horizontal dotted lines show the \textit{ModelOnly} baseline.}
    \label{fig:axis:evolution}
    \Description{Two line plots with x-axis showing the interrogation-synthesis round number from 1 to 10, and y-axis showing the metric score. Correctness scores drop at round 7 but then outperform the baseline. Preference scores drop from the beginning, then stagnate. Goal and action prediction accuracies improve over rounds, except for a sharp drop at round 7, with scores peaking at round 6.}
\end{figure}

Aggregate results for each model are shown in~\cref{tab:axis:eval_results} with per-scenario results and best-round explanations shown in Appendix~G.
AXIS improves the perceived correctness of explanations across all models, by at least 7.67\% and as much as 17.83\%, with improved accuracy on goal prediction from 23.73\% to 46\% for four models, and improved or comparable results on action prediction.
This means that AXIS-generated explanations better capture the causality in the scenario than baselines and are more actionable for downstream tasks.
AXIS also achieves the highest scores for all metrics, although no one model is dominant over others, which indicates that model-level prompt tuning may improve results.

In order to understand how evaluation scores depend on the interrogation-synthesis rounds $N_{\max}$, we evaluate intermediate explanations, plotting aggregate results in~\cref{fig:axis:evolution} and model-level results in Appendix~H.
For perceived correctness and goal/action prediction accuracy, even the intermediate AXIS explanations may improve on the baseline.
In addition, except Llama 3.3-70B, models usually output \texttt{DONE} --- that is, they deemed the available information exhaustive --- before reaching $N_{\max}$ rounds, with average maximum $N=5.03\pm0.27$.
However, LLMs' performance varies over $N_{\max}$, exhibiting biases observed in some previous work, e.g. prematurely proposing solutions, relying too heavily on previous explanation attempts, and a recency bias~\citep{laban2025llmslostmultiturnconversation,chang2024evalsurvey}.
An example of this bias is the drop in scores at $N=7$.
Here, Llama 3.3-70B interrogates counterfactuals completely unrelated to any previous lines of interrogation, e.g. the removal of a previously untested vehicle, and synthesises an explanation based solely on this latest information, ignoring everything that came before that.
This results in a drop in performance, which is mostly recovered in the following round, when the model is re-prompted with the full context.

A similar example of LLM bias can be seen in the lower subjective preference scores for some models using AXIS.
Our evaluation is based on previous work designed originally for human participants~\cite{hoffman2023metrics}, for example, giving a score to the question \textit{``This explanation of why the self-driving car behaved as it did seems complete.''}
However, when we inspect the reasoning of the evaluator LLM, it often gives low scores due to a lack of reference to raw scenario data, for example exact timing or distance information, something that people may consider superfluous~\cite{miller2019socialsciences}.
This also happens across all four axes of the preference metric, and is most pronounced for Llama 3.3-70B, which was instructed to generate concise explanations based on our Shapley analysis.

Finally, some stronger reasoning models (o4-mini, DeepSeek-R1) show significant improvements in perceived correctness, but they have lower goal and action prediction accuracy compared to other models.
This is because reasoning models provide much more context and mention more causes in their explanations, which seem to ``confuse'' the evaluator LLM.
Less advanced models were more succinct in their responses, so the evaluator model could extract the correct goal and next action more easily from the explanation.
This is similar to how people might prefer explanations that are shorter because they might reduce cognitive overload.

\section{Related Work and Discussion}
\label{sec:conclusion}

To answer our research questions, we found that the interrogation-synthesis process of AXIS greatly improves the relevance and correct extraction of causes in rational scenarios, but struggles with irrational behaviour and occluded factors due to the limitations of LLMs.
We also found that AXIS improves the perceived correctness of explanations for all models and actionability for most models, achieving the highest scores across all metrics, suggesting that AXIS is effective at generating useful action explanations.

\subsection{Related Work}
Counterfactual explanation generation has received significant attention in related work but almost exclusively in the single-agent setting, often focusing on mechanistic explanations of neural networks given only a single state/action~\cite{gajcin2024counterfactualsxrl}.
For example, \citet{olson2021counterfactual} developed a deep generative model to produce counterfactual state explanations for RL agents, illustrating minimal changes needed for alternative actions, but their method is limited to explaining a single state instead of full trajectories.
In contrast, action explanation in SDM takes into account the entire trajectory of an agent's actions and their outcomes.
\citet{madumal2020causalxrl} proposed an action influence model grounded in structural causal models to provide counterfactual explanations for model-free RL agents, but their method requires knowing \emph{a priori} which features may have a causal effect.
AXIS instead uses an LLM to automatically propose interventions which may have a causal effect.
\citet{admoni2025SySLLM} introduce SysLLM, which synthesises global textual summaries of single-agent policies by prompting LLMs to describe behavioural patterns based on a dataset of observed trajectories, showing alignment with expert insights, but their evaluation is limited to simple GridWorld environments.
In contrast, we evaluated AXIS using the complex continuous action space of autonomous driving.

In the multi-agent setting, \citet{boggess2022PolicyExplanationsMultiAgent} introduced policy summarisation and query-based language explanations for multi-agent RL (MARL)~\cite{albrecht2024marlBook}, extending this work to tackle temporal user queries~\cite{boggess2023ExplainableMultiagentReinforcement}. %
\citet{heuillet2022CollectiveEXplainableAI} explored the application of Shapley values~\cite{shapley2020value} to cooperative MARL, although limited to explaining single actions.
\citet{gyevnar2024causal} introduced a method which explains actions in MAS using a probabilistic forward model to identify salient causes, performing well in AD evaluations, although their requirement of a probabilistic model is not always easy to satisfy.

\subsection{Limitations and Future Work}

Certain limitations must be considered when interpreting our results. %
\textbf{(1) LLM as external reward model}: While previously \citet{debona2024xaievalllm} have shown that LLM-based substitutes of crowdsourced human data for explanation evaluation are feasible, their results are preliminary, and the assessment of LLMs, especially on subjective preference, is biased, as our results and previous work have shown~\cite{chang2024evalsurvey,anthis2025llmsocialsimulationspromising}.
Future work should focus on running a human subjects evaluation for comparison against LLM-based scores.
\textbf{(2) LLM brittleness in multi-turn conversation}:
AXIS can be thought of as a multi-turn conversation between the LLM and the simulator.
As a result, AXIS is subject to limitations of the underlying LLM, e.g. verbosity, premature ending of conversation, incorrect assumptions, and recency bias~\cite{laban2025multiturnllm}.
We also did not tailor individual prompts for specific LLMs besides our Shapley analysis, and we may see higher scores with better-tuned prompts.
\textbf{(3) Baseline availability and experimental setup}:
As with almost all work in explainable SDM, our results are limited by a lack of comparable baselines from the literature, which we tried to solve by comparing against LLM-only baselines.
Subjective judgements are highly dependent on fluency~\cite{unkelbach2013general}, which further complicates cross-method comparison.
In addition, explanations do not exist without context.
In our case, this meant using the domain of autonomous driving in experiments, and future work should test other domains.
\textbf{(4) Scaling with number of agents}:
We tested AXIS with up to five agents in coupled scenarios with complex continuous action spaces.
However, it is unclear how the method would scale to discrete domains with simpler action spaces where many more agents may be present, e.g. level-based foraging~\cite{christianos2020shared} or warehouse routing~\cite{papoudakis2021benchmarking}.

\bibliographystyle{ACM-Reference-Format}
\bibliography{refs}

\end{document}